\pdfoutput=1

\documentclass[11pt]{article}

\usepackage{acl}

\usepackage{times}
\usepackage{latexsym}

\usepackage[T1]{fontenc}

\usepackage[utf8]{inputenc}

\usepackage{microtype}

\usepackage{enumitem}
\usepackage{graphicx}
\usepackage{transparent}
\usepackage{color}
\usepackage{rotating}
\usepackage[export]{adjustbox}
\usepackage{hyperref}

\title{Handling and Presenting Harmful Text in NLP Research}

\author{Hannah Rose Kirk \\
  University of Oxford /\\
  The Alan Turing Institute \\
  United Kingdom \\
  \texttt{hannah.kirk@oii.ox.ac.uk} \\ \And
  Abeba Birhane \\
  Mozilla Foundation /\\University College Dublin \\
  Ireland \\
  \texttt{abeba@mozillafoundation.org} \\\AND
  Bertie Vidgen \\
  The Alan Turing Institute \\
  United Kingdom \\
  \texttt{bvidgen@turing.ac.uk} \\ \And
  Leon Derczynski \\
  IT University of Copenhagen \\
  Denmark \\
  \texttt{ld@itu.dk}}

\begin{document}
\maketitle
\begin{abstract}
Text data can pose a risk of harm. However, the risks are not fully understood, and how to handle, present, and discuss harmful text in a safe way remains an unresolved issue in the NLP community. We provide an analytical framework categorising harms on three axes: (1) the harm type (e.g., misinformation, hate speech or racial stereotypes); (2) whether a harm is \textit{sought} as a feature of the research design if explicitly studying harmful content (e.g., training a hate speech classifier), versus \textit{unsought} if harmful content is encountered when working on unrelated problems (e.g., language generation or part-of-speech tagging); and (3) who it affects, from people (mis)represented in the data to those handling the data and those publishing on the data. We provide advice for practitioners, with concrete steps for mitigating harm in research and in publication. To assist implementation we introduce \textsc{HarmCheck} -- a documentation standard for handling and presenting harmful text in research.

\end{abstract}

\section{Introduction}
Text data can cause harm in a range of ways. First, different content creates different risks of harm. For instance, misinformation can contaminate the information landscape and polarise groups \cite{mihailidis2017spreadable, au2021role}; Hate speech and abusive language can pollute online communities and inflict long-lasting trauma on its victims \cite{waldron2012harm, vidgen2019challenges}; Negative social stereotypes and misrepresentations of individuals or groups can perpetuate historical injustice and lead to unjust allocation of opportunities or resources \cite{buolamwiniLimitedVision2017, Blodgett2020}. Second, different research settings affect how harms are encountered. Some harmful content is \textit{sought} when researchers deliberately investigate phenomena such as hate speech, extremism or misinformation. In other cases, researchers are working in seemingly unrelated domains (e.g., language generation, part-of-speech tagging or semantic search) but may still encounter \textit{unsought} harmful content, especially if the data are scraped from internet sources~\cite{luccioni2021s,dodge2021documenting,kreutzer2022quality}. 
Third, different groups are harmed by text content during the research process and may suffer immediate, representational or vicarious harms. These groups include \textit{data subjects} (i.e., people represented in the data); \textit{data handlers and researchers}, (i.e., those who collect, annotate or audit the data, and produce research outputs)~\cite{pyevich2003relationship, vidgen2019challenges,newton2020facebook}; and \textit{readers and reviewers} (i.e., those who read research outputs).

Each of these complexities -- \textit{what} the source of harm is, \textit{how} one is harmed and \textit{who} is harmed -- presents concerning ethical and methodological challenges that need to be addressed for the NLP field to advance in a responsible and equitable manner. If left unaddressed, inadequate safeguarding can
stifle healthy research practice around harmful content. Careless presentation of examples in academic papers perpetuates negative portrayals of data subjects and could distress readers, while incautious research protocols inflict an emotional toll on data handlers, annotators and researchers. Furthermore, if not explicitly stated, researchers risk being misconstrued as aligning with the views expressed in harmful content. Despite these concerns, avoiding any research that bears a risk of harm is untenable because harmful content is a feature of many datasets (and the real world). Bringing text harms to light and communcating their features via research is a critical first step to tackling them.
\newpage
We address these unresolved challenges, and provide both an analytical framework that describes the sources of harm, who they affect and how; and a set of guidelines which outline the responsibilities that researchers have, to others and to themselves, when handling and presenting harmful text. 
To ease adoption of our recommendations, we present \textsc{HarmCheck} -- a checklist for transparent, responsible, and reflective reporting of text harms in research. While algorithm audits and the risks of harm arising from model deployment have received relatively widespread attention \cite[for a summary, see][]{weidinger2021ethical}, there is significantly less research documenting the risks that arise pre-deployment, i.e., during the research design, dataset processing and publication stages. The novel contribution of our work is to establish a common ground and encourage better protocols for continuing empirical research on harmful text in a safe and proportionate manner.
Note: any paper that presents harmful content should include a clear content warning in the introduction or abstract at least a page before any examples are shown.

\textcolor{red}{\textbf{Content Warning:} \textit{This document discusses examples of harmful content (hate, abuse, misinformation and negative stereotypes). The authors do not support the use of harmful language, nor any of the harmful representations quoted below.}}

\section{Harms and Risks in Text Data}
\subsection{What is Harmful Content?}
We use the term `harm' 
to refer to both the content which creates a risk of harm (i.e., hate speech as a \textit{cause} of harm), as well as the negative impact of that content on the emotional, psychological and physical well-being and safety of individuals, groups or society (i.e., psychological harm as an \textit{effect} of hate speech). In practice, whether a form of content inflicts harm (and, if so, the degree of harm that it inflicts) depends on a range of intersecting factors, including the nature of the content; the immediate and broader context; the historical setting; where it comes from; who it is directed at; and who encounters it. 
Small differences in these factors can make a substantial difference, and not all content that presents a risk of harm will actually inflict harm in every case. 
In this paper, we adopt the position that `harmful content' both constitutes a harm in-of-itself (i.e., it is harmful because of its intrinsic features) and causes harm because of the substantial risk of detrimental effects on individuals, groups or societies (i.e., it is harmful because of its impact) \cite{waldron2012harm}. 

Previous works have proposed taxonomies for understanding various risks \cite{weidinger2021ethical} and types of harmful content \cite{bankoUnifiedTaxonomyHarmful2020}. However, types of harmful content and the associated risks are, by their nature, complex, emergent, contextual and changing -- they are an `open class' problem, 
which means that enumerating all varieties is 
not possible. 
We focus primarily on harmful content that is associated with discrimination, exclusion and toxicity, as well as informational harms \cite{weidinger2021ethical}. For example, our scope includes misinformation~\cite{derczynski2015pheme}, propaganda~\cite{da2020semeval}, incendiary and manipulative messages, descriptions of harmful acts, hate speech,  abuse, slurs and threats of violence~\cite{vidgen2020directions}, as well as sexist, racist and otherwise marginalising or negative stereotypes~\cite{birhane2021large}. Other forms of illegal content, such as child abuse and terrorism material, present risks that could be covered by the guidelines here but likely require supplementary safeguarding given their severity. 

The nature of the harm that individuals experience from harmful content varies considerably. It can depend on duration 
(either \textit{short-} or \textit{long-term}); and how it manifests (either \textit{internalised}, such as suffering negative emotions, or \textit{externalised}, such as exclusion from resources due to misrepresentation). 
Externalised short-term harms may be more obvious but internalised and long-term harms can be just as damaging. For instance, online hate can severely damage the mental health of victims -- a \textit{long-term} and \textit{internalised} form of harm \cite{gelberEvidencingHarms2016a}. 
Harmful text can inflict negative effects on society, as well as individuals. These harms can be harder to identify because they are pernicious and nuanced, creating wider and more diffuse negative effects. Individual- and societal-levels harms often interact: in one direction, societal-level harms can deepen individual-level harms by increasing the affective pull of harmful content. Representational harms, for example, emerge from sexist, racist, ableist, and otherwise unjust historical, cultural and societal norms, which are then embedded in data \cite{Blodgett2020,ahmed2007phenomenology}. Such harms can lead directly or indirectly to allocational harms, where under-served groups face inequitable constraints to resources, reflecting back a deep-rooted culture of injustice and power asymmetry. In the other direction, individual-level harms can accumulate into societal-level harms. For example, electoral misinformation can lead to individuals attending the wrong location to vote, disrupting the democratic process; while climate change or vaccine misinformation targeted at individuals can place a negative externality on wider society.

The designation of content as harmful has social, political and methodological implications. It is often difficult to quantify and measure the degree of harm inflicted on individuals, groups or society because
the materialisation of harms is intimately related to identity, lived experience and context. The same content could affect individuals idiosyncratically: 
in a study of perceptions of hate speech, \citet{costello2019social} show that men and political conservatives find hateful material less disturbing than women or liberals, while \citet{sapAnnotatorsAttitudesHow2022a} find annotators who scored highly on their scale for racist and conservative beliefs rated anti-Black language as less toxic.
This raises the question of \textit{who decides what is harmful?} Critical data scholars~\cite{benjamin2019race,d2020data,birhane2021algorithmic} contend that those at the receiving end of harm and injustice hold the epistemic prerogative to define harm due to their lived experience -- while those occupying the most 
privileged positions in society are poorly equipped to recognise it, a phenomena that~\citet{d2020data} term the \textit{privilege hazard}. For example, given the problematic history of the term, it should not be up to white researchers to decide whether the use of ``\textit{n***a}'' is offensive or not. The experiences of individuals and communities at the margins of society who often disproportionately face abuse, hate speech and marginalisation must direct and shape understanding of harm~\cite{benjamin2019race}. 
Thus, while we outline some of the key considerations in handling and presenting harmful content, we recognise that assessing harm is always a reflective, contextual and ongoing process.

\subsection{How is Harm Encountered?}
Text data is fundamental to nearly all areas of language processing, and good quality data 
is critical for better performing, safer and more robust models~\cite{sambasivan2021everyone, xu2021dataclue}. The shift towards widespread use of pre-trained models has intensified the search for more and higher quality data within the NLP community. To deal with the demands of deep learning, many data curators and researchers have turned to internet-scraped datasets, such as the Common Crawl Corpus or WebText \cite{radfordLanguageModels2019}. Large-scale language models pre-trained on such datasets have become popular because of their high performance and the ease with which they can be accessed, such as using Hugging~Face Transformers~\cite{wolf2020transformers}. They are now commonly-used `out-of-the-box' for a variety of downstream tasks~\cite{kirk2021bias}. Most researchers can only fine-tune these models on small datasets because retraining a large model from scratch is infeasibly resource-intensive.   Recent work has prioritised data acquisition over model complexity~\cite{paullada2021data} and forefronted the importance of data audits~\cite{koch2021reduced}. While these developments are welcome, it may mean that researchers spend more time qualitatively inspecting datasets, auditing their content, and manually reviewing labels. This creates new venues for harm. However, exactly how harms transpire during data auditing 
and model training depends on whether harmful content is \textit{unsought} i.e., encountered unexpectedly in the dataset, versus \textit{sought}, i.e., collected as a feature of the research design to investigate harmful phenomena. 

\paragraph{Unsought Harmful Content} The potential for encountering harmful content exists even when one is not looking for it, especially with the wider use of large-scale, internet-scraped datasets. The enormity of these datasets increases the likelihood of harmful content and decreases the tenability of human audit for quality or toxicity~\cite{hanna2020against,luccioni2021s,kreutzer2022quality}. Harms in datasets can transfer to the models trained on them~\cite{10.1145/3442188.3445922b, rauhCharacteristicsHarmful2022}. Large language models have emergent capabilities which are difficult to fully understand \cite{bommasani2021opportunities}, creating a risk that harms will be inadvertently propagated in novel and unexpected ways from derived model's behaviours, generations or predictions~\cite{gehman-etal-2020-realtoxicityprompts}. We term this latent risk  \textit{unsought} harm. It is particularly noteworthy because people may not be fully aware of it, and so have an increased chance of not preparing for it. 

\paragraph{Sought Harmful Content} In some research, harmful content is actively sought out e.g., when compiling a dataset of hate speech, or auditing a dataset for toxic content. 
In this case, data curators, auditors and researchers 
typically cannot avoid coming into contact with harmful content -- it is difficult to study hate speech without a dataset that contains hate. 
When harmful content is the focus of a study, the researcher expects to encounter, and may even target, harmful text. We term this \textit{sought} harmful content. On the one hand the effect of \textit{sought} harmful content poses a greater risk due to the increased density of harmful content. Yet on the other hand, researchers, auditors, annotators and reviewers are somewhat aware of the risks \textit{a priori} and can prepare accordingly~(see \S\ref{sec:handling}).

\subsection{Who is at Risk of Harm?}
Harms in text data can affect many groups. We identify three (non-mutually exclusive) risk groups.

\paragraph{Those represented in the data} Humans represented in the dataset are at risk of harm from what it contains and what it omits. Harms arise from `hyper-visibility'~\cite{noble2013google}, where excessive data portraying negative attributes such as  
stereotypes is collected ~\cite{birhane2021large}. 
Harms also come from `hypo-visibility', a form of erasure whereby the lived experiences of entire groups and communities are omitted from the data, thus rendered invisible to NLP systems~\cite{jo2020lessons}.
These two directions of harm, hyper- and hypo-visibility, can exacerbate each other if (a) certain groups are represented rarely and (b) those representations are exaggerated harmful portrayals. 

Presenting harmful content in research publications without the necessary precautions and safeguards can propagate the harm to data subjects. 
This is the case for misinformation where spreading known-harmful ideas and false claims without making the problems with them unavoidably evident can perpetuate the harm that is caused. For instance, exposure to false headlines increases the chance of their claims being accepted and normalised, even when the reader knows they are false~\cite{pennycook2018prior}. 
Research that presents negative stereotypes without contextualisation and qualification also risks further entrenching the associations in the dataset, deepening the harm to the data subjects~\cite{barlas2021see}. In particular, when researchers are from a different background to those that are subject to harms, there is a greater risk of `dehumanisation-by-datafication', where the content is treated as de-humanised data studied in an abstract manner -- rather than something that has direct implications for the subjects' representation, welfare, and safety~\cite{leurs201715}. 

\paragraph{Those working with the data} 
People who work with harmful text at any stage during the research process are at risk of vicarious trauma through their exposure to harmful content, particularly if it is repeated~\cite{pyevich2003relationship,dubberley2015making,newton2020facebook, steiger2021psychological, spangenberg}.
\textit{Dataset curators} are exposed to harm when collecting dataset entries; for example when using keyword searches on a social media platform's API to find content. After data is collected, \textit{data processors or engineers} are exposed when writing code or cleaning data; for example, inspecting datasets that contain a high-proportion of abuse. Exposure to harms also arises during analysis; for example, in unsupervised learning, topic labels are assigned by reading the most representative documents, or in supervised learning, entries are given labels and models may be interrogated with qualitative error analyses. \textit{Data labellers} or \textit{annotators} are at particular risk of harm as they are burdened with assigning labels to data entries. Data labelling and annotation is a critical yet undervalued part of NLP research and data labellers, as they deal with fine-grained details of data, are at high risk of harm. 
During the publication stage, there is a welfare risk to \textit{authors} as they discuss, summarise or directly quote examples; as well as a reputational risk if these examples are misconstrued as representing author beliefs through careless reading, ambiguous presentation, or being taken out-of-context.

\paragraph{Those consuming research about the data}
People reviewing and reading papers or attending talks produced about the data, may be distressed by exposure to harmful content. If such examples are not properly justified then reviewers can object to their inclusion, give negative reviews or propose desk rejection. Consider this example (synthetic) review comment on a paper investigating hate speech:

\begin{quote}
     Ethical issue: Even though the authors added a trigger warning in the paper, it was still uncomfortable for me to see examples along the lines of \textit{``I want to murder Muslims"} in this manuscript. Researchers should confine themselves to discussion of their novel methods; it's not relevant to include so many distracting and useless quotes.
 \end{quote}
  \vspace*{-50mm} \hspace{11mm} \hbox{%
   \bfseries\begin{turn}{45}
   \huge \color{red} \transparent{0.25}
   HARMFUL QUOTE
   \end{turn}
  }

The risk to readers' welfare must be balanced with the need to provide examples that demonstrate the nature and severity of the research undertaken. As such, harmful examples should always be clearly justified and carefully presented.

\section{Guidelines for Handling and Presenting Harmful Text}
In areas of academic research outside of NLP, there are well-established practices for reducing the risk and severity of harm that researchers experience.  
People working in a chemistry lab are protected by safety protocols for working with hazardous materials, such as poisonous gases or radioactive substances.
Analogously, there should be protocols for reducing the risk of harm from hazardous materials in language. This section outlines ways to reduce the risk of harm from text data. We present our guidance in chronological order of the contact points that arise during research and practice.

\subsection{Mitigating Harmful Text from the Start}
\label{sec:handling}
The first opportunity to mitigate harm is during data curation and selection. If harms to data subjects are not tackled at this stage then they can become `frozen in time' within the dataset.  
When harmful content is \textit{unsought} because the reason for creating the dataset is for other NLP tasks (e.g., training a large language model), the onus lies on researchers, data creators and curators to audit and safeguard their datasets prior to public release. 
Indeed, some unlabelled corpora used to train large language models are filtered to remove the most obvious forms of toxic language \cite{brown2020language}. 
Filtering, while well-motivated, must be cautiously approached because it can censor and erase marginalised experiences. For example, crudely removing language that could be considered offensive (e.g., \textit{any} use of potentially reclaimed terms, such as ``\textit{sex}'' or ``\textit{gay}'') risks excluding the language of entire communities who may use such terms to communicate about sexual health~\cite{dodge2021documenting}. 

Harmful content that is 
\textit{unsought} can potentially be as harmful as \textit{sought} harmful content. 
However, there is substantially more literature and audit work documenting this \textit{unsought} form of dataset harm \cite{luccioni2021s, DBLP:journals/corr/abs-2110-01963, dodge2021documenting}.
Accordingly, researchers, data auditors and annotators that actively engage with \textit{sought} harmful content are 
the main focus of our guidelines for handling, presenting and publishing research.

\subsection{Handling Harmful Text During Research} 
Several practical steps can be taken when handling harmful text.

\paragraph{Brief} 
It is important that the 
objectives of the research are well-understood, as well as the likely risks that will be encountered. Working in teams (as opposed to solo) helps distribute the amount of harmful text one is likely to be exposed to. 
Research teams should avoid engaging in projects without at least one researcher who has some prior experience or without extensively reviewing prior research and critically examining the upcoming task. 
Researchers need to make a realistic and well-grounded assessment of the risks that they and others are likely to face \textit{before} starting work. 

\paragraph{Check-in}
Check-ins between team members should be regular and frequent, and there should be a direct channel of communication between all researchers (including annotators). Senior researchers should provide feedback mechanisms that are both anonymous and individual, giving opportunities for people with different communicative preferences and needs to provide meaningful updates on their experience of the work. At a minimum, there should be space for people working with harmful text to talk about their experiences, even if just to share anecdotes or vent~\cite{riskyresearch}. 
Regular feedback can also aid the research process by creating multiple touchpoints between all parts of the research team. This could affect research design as, for example, sustaining communication may be more complex with third-party crowdsourced workers. 

\paragraph{Limit} 
The risk of researchers experiencing harm can be minimised by reducing their exposure to content. For some researchers, such as annotators, exposure is unavoidable -- but can be limited with more efficient techniques for working with data. For instance, active learning and transfer learning reduce the total amount of labelled data needed for a given project. In some fields, such as computer vision, techniques have been developed to enable annotators to carry out their work whilst minimising the risk of harm, such as greyscaling or blurring images ~\cite{karunakaran2019testing, das2020fast}. Similar approaches for text could be considered, such as masking harmful words, although this may constrain the actual work of annotation. For people involved in other parts of the research process, engagement with data can be substantially minimised by more effective data processing. For instance, harmful text can be replaced with dummy data for establishing coding pipelines and the real data only used once models need to be trained. 

\paragraph{Support}
Mental health and psychological support services should be in place for those who work with harmful text. 
Providing support can help to address negative experiences when they occur, and develop coping mechanisms 
\cite{steiger2021psychological}. 
Resources should be varied and practically focused, and fit the needs of the person at risk of harm. In-house support and counselling services may need to be paid for but could be infeasibly expensive for the average research lab. In these cases, researchers may be referred to campus- or company-wide services if avaliable, or at least, be pointed towards publicly-avaliable resources.\footnote{For example, the Vicarious Trauma Toolkit compiles a list of 500 freely avaliable resources, \url{https://ovc.ojp.gov/program/vtt/compendium-resources}.}
Providing support is particularly important when there is social stigma associated with seeking help or where those working with data are concerned about how they are perceived. 

\paragraph{De-brief} 
At the end of the research, coordinators should explain to the team the anticipated impact of the research on harmful content, and discuss any unique or unanticipated issues that were encountered. 
This process should be as `horizontal' as possible, enabling all researchers to express their views and experiences in an open dialogue. 
The de-brief is a useful opportunity for researchers to identify lessons learnt, refine processes and take steps to mitigate the risk of harm in the future, further building team resilience and potentially improving the quality of future research.

\subsection{Presenting Harmful Text for Publication}
When publishing research about harmful phenomena, authors need to take steps (1) to protect and respect those represented in the dataset, (2) to warn of harm and limit exposure to those reading the research, and (3) to distance their own opinions from the harmful views or examples being discussed. These aims can be achieved using a selection of techniques: preview -- distance -- disclaim -- respect, inspired by journalistic practice~\cite{njfact,factchecking}. While the best way of reducing risks of harm is to not give examples of harmful content at all, precise exposition and argumentation of a method or motivating problem in research sometimes requires these examples. If harmful content is quoted in research outputs, it should be presented carefully, consistently and respectfully.

\paragraph{Preview} Readers need to know what to expect. Authors should preview or signpost harmful content in a consistent fashion. They should avoid placing harmful content on the first page or above the fold, so that the audience gets a chance to decide whether they want to see it. A clear and visually distinct `content warning' ahead of potentially troubling content can be useful. While it has been noted that `trigger warnings' can risk reinforcing harm~\cite{bridgland2021danger}, the benefits from transparent signposting of harmful content likely outweigh the risks.

\begin{figure}[h]
    \centering
    \includegraphics[width=\columnwidth, frame]{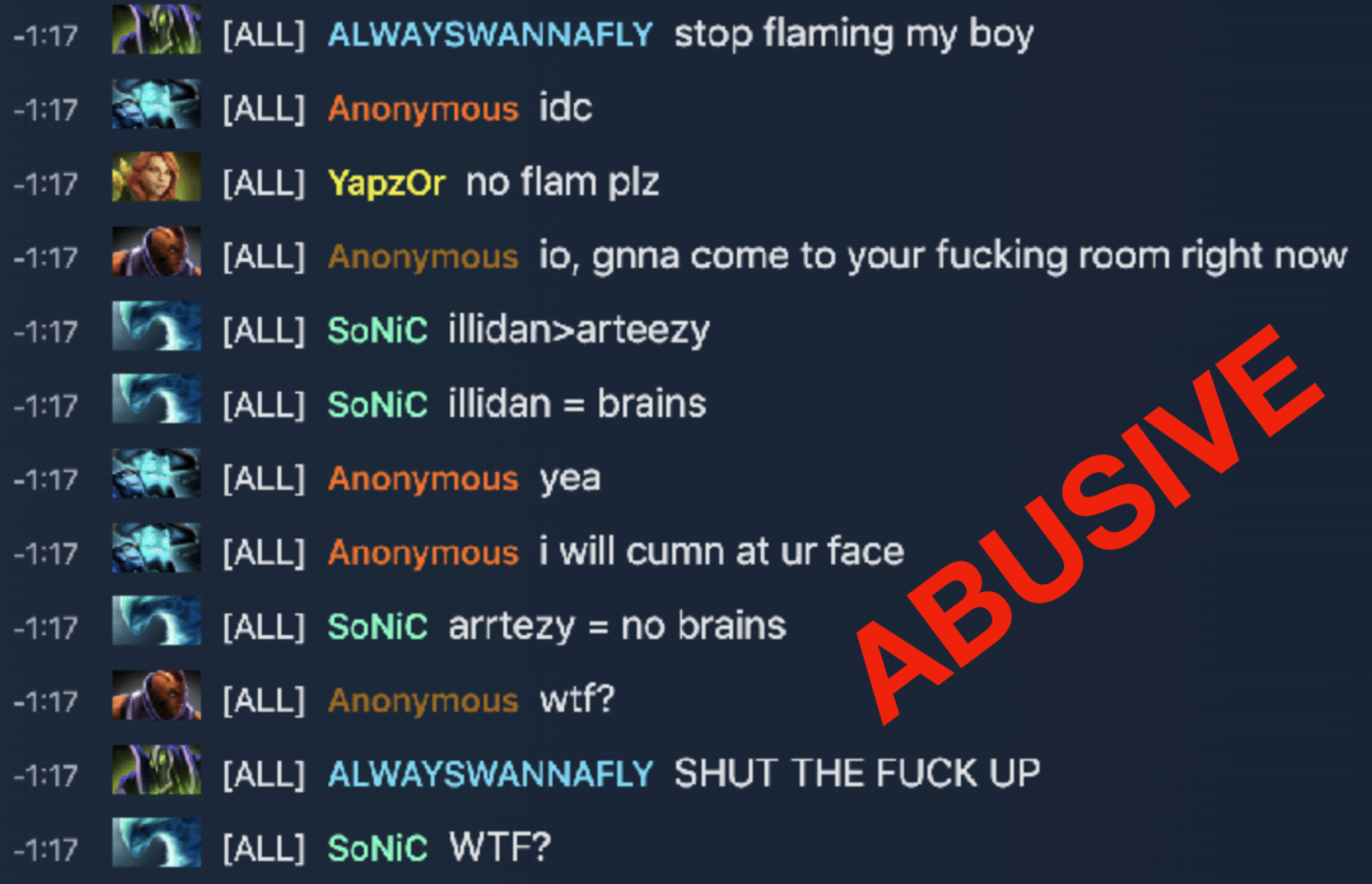}
    \caption{Example of an abusive exchange 
    from an online game.}
    \label{fig:dota2}
\end{figure}

\paragraph{Distance} In the case of harmful content, it is important to clearly distance the research \textit{on the data} from the viewpoints and material contained \textit{in the data}. Using minimal examples distances the research output from the data itself; but if examples need to be used, they should be formatted consistently so that it is clear to even a casual observer that an example of inciteful, biased, false or hateful content is not written by the authors themselves. This can be achieved visually, by including a bold highlight by each problematic example or including a watermark that overlaps the example in the paper (see Figure~\ref{fig:dota2}).
Harmful examples can also be presented less strongly, for example by reducing contrast in an in-line text example with grey font as an NLP analogy to work on blurring and greyscaling images~\cite{karunakaran2019testing, das2020fast}. Replacing identifiers or slurs with placeholders, e.g., \textit{``that [IDENTITY] is a [SLUR]"} or \textit{``I hate [IDENTITY]"}, can convey syntax and some semantics without reproducing harm targeted at a specific target group \cite{rottger2021hatecheck, kirk2021hatemoji}.

An addendum to this guidance point is that some content may be worth explicitly \textbf{disclaiming}.
This is particularly relevant for misinformation research, where examples of false claims should be explicitly flagged and accompanied by the relevant true claim. 
For example, it is worth stating alongside a quote expressing a false narrative about the MMR vaccine and autistic spectrum disorders that the vaccine is not connected to these results, and including a reference, so that the message is clear. If possible, the provenance of the content should be identified, provided this is in-line with privacy regulations and ethical concerns; it sometimes may be best to say the platform of origin and date e.g., ``From a Twitter user, November 2020".

\paragraph{Respect} 
It is critical that the groups represented in, and often targeted by, harmful content are treated with
respect. Researchers should engage in active and continual reflective practices, such as striving to adopt the perspective of those represented in data and with the aim of developing critical awareness of the social and historical roots of groups and identities that are subject to harm. This might entail interrogating any concepts, terms or phrases used in data to describe or represent identities in a way that adheres to social stereotypes, then challenging or countering such conceptions. Respecting data subjects also entails protecting their dignity and privacy, for example by removing identifying information (such as Twitter handles), by making sure groups or identities are not described using pejorative terms (if this has been the case with the original data) or by blurring any identifying images.

\subsection{Preparing to Publish Research on Harmful Text}
As with any research, researchers working on harmful content need to publicise and
disseminate their work. 
However, this comes with its own risks. Researchers have faced attacks, both online and offline~\cite{riskyresearch, vogels2021state}, and have been subjected to online abuse, death threats, deepfake revenge porn, doxxing (finding and publishing personal information) and even swatting (in the USA, having an armed unit storm the researcher's house with guns)~\cite{mortensen2018anger,greyson2018online}. For example, one paper on gender bias published at NeurIPS \cite{kirk2021bias} prompted a wave of misogynistic attacks against the lead author on Twitter.
Another paper on online misogyny published at ACL \cite{zeinert2021annotating} prompted large amounts of online abuse and doxxing directed at the authors by name; frivolous freedom of information requests explicitly for the purpose of wasting time; complaints made to the authors' external funding organisations; public attacks from politicians against the authors and their institution; and pejorative opinion articles in the national press against the research. 
Although these interactions are completely unacceptable, they 
to some extent form part of many researchers' experience of working in areas related to harmful content. We present these stories not to deter research in the NLP community on harmful text, but instead to encourage safeguarding practices: when publishing research in these high-risk areas, authors should take concrete steps to reduce risks and to protect themselves. 
Note that the abuse that researchers are often subjected to when studying harmful content, which 
is not the norm in other fields, goes far beyond legitimate academic discourse. In that spirit, our suggestions are not intended to constrain academic and civic discussion about research -- it is certainly the case that some criticism of research outputs in risky areas will be legitimate, even if heated. Proper documentation of research outputs (such as appropriately labelling harmful content in datasets) actually increases research transparency and aids better discussion. 

\paragraph{Inform} A researcher can give their organisation (and its press, communications and legal departments) advance warning that they are publicising the research, and that it may bring some harassment. Organisations should have procedures for handling this and protecting their members~\cite{ketchum2020report}. If no procedures are in place, guidance and policy templates are publicly available for researchers to initiate dialogue.\footnote{E.g., Data \& Society's sheet on \href{https://datasociety.net/wp-content/uploads/2016/10/Online\_Harassment\_Information\_Sheet-Oct-2016.pdf}{`Online Harassment Information for Universities'}} Informally, finding people to discuss experiences of harassment with, either professionally or as a friend, can build a support network in case of a backlash.

\paragraph{Protect yourself} Would-be harassers may search online to find details about their targets or to identify routes of attack (e.g., via abusive social media messages). Researchers may wish to consider editing, hiding, or removing online information that they would not want malicious parties to use, such as personal emails or phone numbers \cite{glaserprotect}.

\paragraph{Curate outreach} Talking to the press is rarely compulsory: not every media request has to be answered. Some discussions are likely to lead to negative coverage. It is worth examining what a journalist or outlet has published before in order to make an informed decision about whether or not to engage.










\section{\textsc{HarmCheck}: A Checklist for Handling and Presenting Harmful Text}
In recent years, there has been a movement towards the responsible and transparent documentation of research artefacts \cite{bender2018data, mitchell2019model, rogers2021just}, as well as encouraging best practices for ethical NLP research \cite{leidnerEthicalDesignEthics2017, smileySayRightThing2017}. Some conferences now require that authors fill in a responsible NLP checklist to accompany their submission.\footnote{\url{https://github.com/acl-org/responsibleNLPresearch}} 
In a similar vein, we 
encourage standardised documentation of harmful content contained in research outputs. 
To this end, we present \textsc{HarmCheck}, a simple checklist that works as both a standalone piece of documentation and as an addendum to existing documentation standards. It is intended for people specifically researching \textit{sought} harmful content. 
Each section is designed to be filled in as a statement (such as in a data statement~\cite{bender2018data} or model card~\cite{mitchell2019model}), with some sections being more appropriate for different harm types. Researchers should focus on the areas that are most relevant to their research. 

\textsc{HarmCheck} should not be seen as a compliance form, but as a starting point for a broader conversation about the risk of harm through research. \textsc{HarmCheck} is not exhaustive of all possible harmful text or ways of handling them. Transparency through documentation should be seen primarily as an opportunity for reflecting on and scrutinising research, rather than a box-ticking exercise. Although we provide a list of starter questions for each section to guide researchers, they should not be seen as a definitive or complete set for assessing harm in all cases. 
In Appendix~\ref{sec:harm_checK_examples}, we demonstrate how \textsc{HarmCheck} can be used by providing completed harm statement examples for three types of harmful text: hate speech, misinformation and negative stereotypes.

\subsection{Proposed Checklist}

    \paragraph{Risk of harm protocol} Summarise any steps taken during the research progress to identify and mitigate harm to at-risk groups.
    \begin{itemize}
        \item What are the specific risks of harm and to who? Have you 
        outlined how the well-being of any researchers, annotators or data processors was protected during the study period?
    \end{itemize}
    
    \paragraph{Preview} Summarise all content warnings for harmful content.
    \begin{itemize}
        \item Is there a content warning at least a page before any harmful text instances are presented? Is the content warning clearly visible? Do section, table or figure- specific content warnings describe the nature of the harm?
    \end{itemize}
    
    \paragraph{Distance} Summarise the steps taken to (a) clearly identify which parts of the text are harmful examples and (b) separate the author(s) from the content.
    \begin{itemize} [noitemsep]
    \item Is it clear that the harmful text is not part of the material's body? Is there visual distinction of harmful examples with a watermark or text colour? Are harmful examples given reduced prominence relative to the containing document?
    \end{itemize}
    
    \paragraph{Disclaim} [\textit{optional and most relevant for informational harms}] Summarise any corrections or counter-claims to harmful content or documentation of its source.
    \begin{itemize}
    \item Is the origin of the harmful text clearly identified? Are harmful and false claims explicitly disclaimed or corrected? 
    \end{itemize}
    
    \paragraph{Respect} Summarise steps taken to interrogate potentially marginalising and pejorative terminology or framing of groups and identities represented in data, as well as measures taken to protect their dignity and privacy. 
    \begin{itemize} 
    \item Have harmful words, slurs or phrases targeted at data subjects been minimised where possible? Has personally-identifying information, images or text been removed, blurred or anonymized? 
    \end{itemize}

\section{Conclusion}
Harmful content can be contained in datasets, and 
encountered whether one is looking for it or not. Various forms of harmful text risk inflicting serious negative effects on many groups of people -- 
prior to, during and after the research process. Some professional and academic areas have established protocols for dealing with their inherent hazards and harms -- and we have now proposed such a protocol for NLP research. 

The harm caused by textual content does not emerge in a social, cultural or historical vacuum -- data embeds and perpetuates social norms, historical injustices, and uneven power dynamics. As such, technical fixes from individual researchers or adopted protocols cannot fully address this problem. Instead, it requires acknowledging and contesting unjust systems and their manifestations in text data, envisioning what worldview the data should (or should not) represent, and eventually working towards making such visions a reality. We do not suggest that the NLP community can alone bear the weight of responsibility for countering the deep-rooted historical, cultural and societal issues in-grained in language data -- this remains an unresolved problem which requires systemic change from multidisciplinary perspectives. At the same time, while tackling harmful content in datasets requires broader systemic change, adopting better practices in discussion and presentation can incrementally contribute to such change. 

Thus, another unresolved problem in NLP is how the production, sharing and consumption of research itself can be approached in a way that encourages awareness and mitigation of harms. This paper has identified the problems and risks that emerge when handling and presenting harmful text. It also provides practical solutions, including the introduction of \textsc{HarmCheck}. Discussions about the harm caused by and through NLP research, as well as other computational and data-intensive areas, are still at an early stage, despite the progress made in recent times.
As such, the arguments and recommendations we have presented here are only the start of a larger conversation about risks and harms in our field. We hope that delineating and describing them opens a broad dialogue in the NLP community towards 
responsible, safe and ethical research on harmful text.

\section*{Limitations}

\paragraph{Context} Our work is focused on the setting of conducting NLP research. However, many people process text, sometimes even containing \textit{sought} harms. The practices and norms in their contexts may be different from those highlighted here, and we have not conducted a survey of the broad range of ways in which people interact with text data. Instead, we have assumed a few typical workflows that the authors are aware of, and built recommendations based on these. To address this limitation, our focus has been placed on using broad terminology and suggesting multiple scenarios when giving examples or considering risks and impacts.

\paragraph{Harms are an open class} As stated in our work, harms are an open class: the kinds of harmful text that datasets might contain is inexhaustible. While we give examples concerning a few of these, it is almost certain that there remain additional kinds of harms and risks in text that we have not mentioned or addressed in our recommendations. To this end, we have preferred broad terminology and clearly defined the scope of our research. Our scope primarily focuses on harms arising during the dataset curation, research process and publication stage -- we do not comment on downstream harms caused by models trained on datasets containing harmful text.

\paragraph{People are an open class} What can be considered harmful and its effects varies vastly from person to person. Many of our recommendations may not be useful in some contexts; we hope that most aren't needed at all. However, text that is harmful to any group, \textit{is} harmful text. To this end, we 
have based our examples and practices on established cases of text harm that 
demonstrable negative impacts on data subjects, researchers and readers, while recognising that this is not an exhaustive list of either harmful text or impacted groups. 

\paragraph{Positionality}
In presenting this work, part of our objective is to minimise harm for all involved in the research process including data labellers, auditors, model builders, researchers, academic reviewers and the general public that might come into contact with published work. Even though we have aspired to approach the question of harm in a way that is inclusive of perspectives from marginalised groups, we acknowledge that by virtue of our positionalities (we are all well-resourced researchers based in Western universities and institutes), we inevitably have blind-spots. For instance, we cannot contend to have centred the perspectives from disfranchised, often underpaid, and vicariously employed data labellers or annotators. We consider (and encourage others to consider) these workers as part of the research team so strive to be inclusive of their perspectives, hoping that such reflective practice serves as a critical initial step.  

\bibliographystyle{acl_natbib}
\bibliography{custom}

\appendix
\clearpage
\section{\textsc{HarmCheck} Case Studies}
\label{sec:harm_checK_examples}

This section contains applications of \textsc{HarmCheck} to three pieces of prior work, each addressing a different kind of harmful text. The examples detail what is done or what could be done.

\subsection{Multimodal Negative Stereotypes}

\begin{figure}[h!]
    \centering
    \includegraphics[width=\columnwidth, frame]{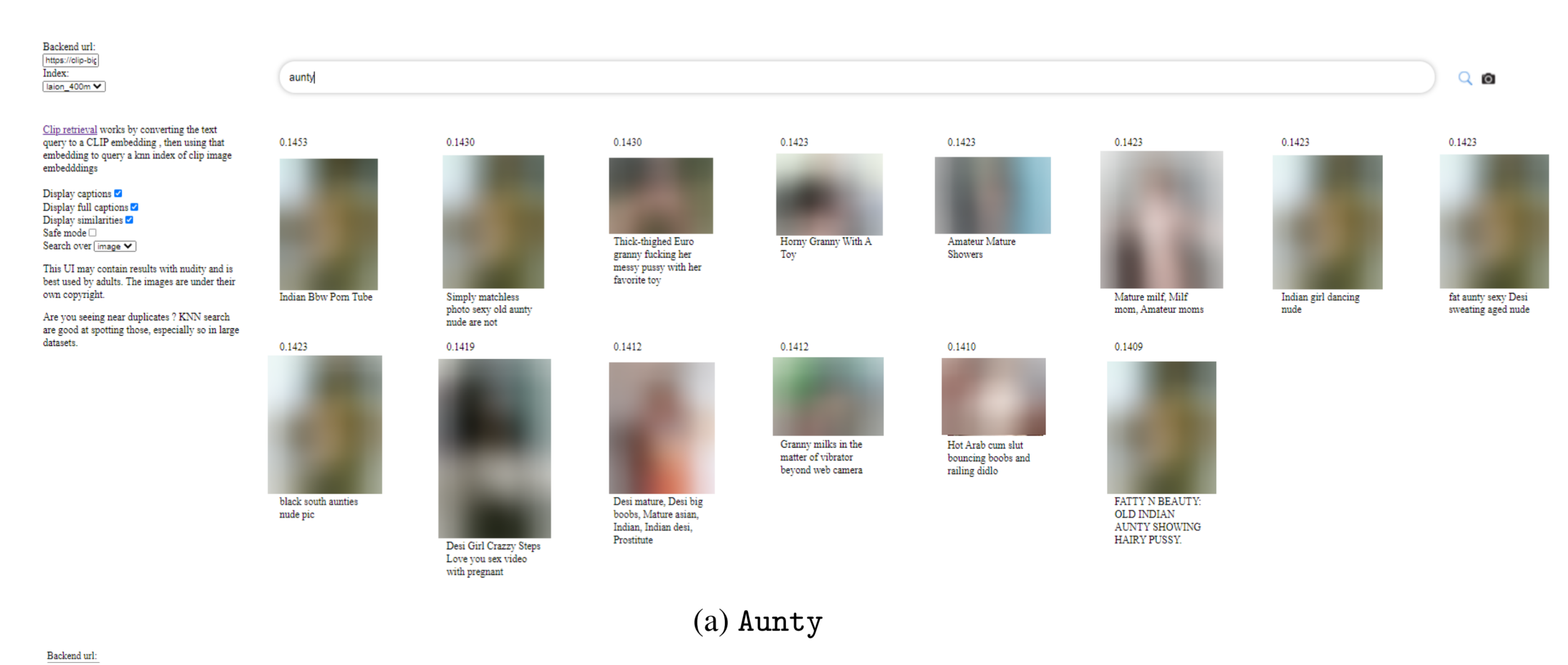}
    \caption{Screenshot from~\citet{DBLP:journals/corr/abs-2110-01963} where the authors present a result showing bias. Note blurring to avoid re-associating image subjects with the harmful biases.}
    \label{fig:laion-bias}
\end{figure}

\citet{DBLP:journals/corr/abs-2110-01963} examine an automatically-generated multimodal dataset and find that the dataset contains explicit image-and-text pairs of rape, pornography, malign stereotypes, and racist and ethnic slurs. The paper outlines steps that dataset creators could take to improve the dataset.
\begin{enumerate}
\item \textbf{Risk of harm protocol} There is a risk of harms to data subjects from the propagation of harmful stereotypes and association of the people in the images with derogatory and inaccurate terms. There is a risk of harm to researchers from sifting through the dataset for a prolonged time period. The study included scheduling regular check-ins and authors took turns at handling the most problematic aspects of the data. There is a risk of exposing harm to those sharing the work environment such as family and colleagues. The research was conducted in isolation (not in a lab environment) to reduce the risk of others accidentally seeing the screen while it contains troubling content. There is also a risk of harm to the reader from exposure to explicit, distressing and/or offensive images. 
\item \textbf{Preview} A content warning is included directly after the abstract which describes the nature of the content and is positioned at least a page before harmful content is shown. The warning is italicised for visual impact. At the beginning of Section 2, there is a further note that all offensive imagery from this section has been blurred and moved to the Appendix after a blank page to give the reader the choice to not visually engage.
\item \textbf{Distance} The paper avoids propagating harms by indicating clearly that the content is an example -- achieved through a different typeface, by stating it explicitly, and by using the same presentation style for examples throughout the paper. 
\item \textbf{Disclaim} In order to clearly indicate that the images came from the dataset portal, screenshots (which were then blurred) included the portal interface.
\item \textbf{Respect:} Images are blurred to protect subjects' identity and remove the visual association between image and text for the reader (see Figure \ref{fig:laion-bias}). Following a series of reflective discussions amongst the team, the authors also took the necessary precautions to avoid the framing and portrayal of data subjects in terms and phrases that adhere to social stereotypes.
\end{enumerate}

\subsection{Emoji-Based Hate}
\begin{figure}
    \centering
    \includegraphics[width=\columnwidth, frame]{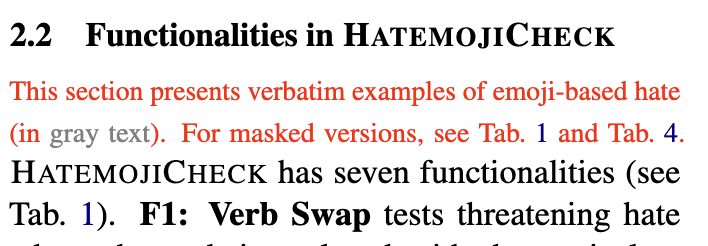}
    \caption{Screenshot from~\citet{kirk2021hatemoji} where the authors add a section specific warning. The content warning is in \textcolor{red}{red} for visibility and hateful examples in \textcolor{gray}{gray} to displace their prominence.}
    \label{fig:hmoji_sec_warning}
\end{figure}

\citet{kirk2021hatemoji} study emoji-based hate, and provide two datasets for better evaluation of model vulnerabilities and better training of classifiers.
\begin{enumerate}
\item \textbf{Risk of harm protocol} There is a risk of harm to data subjects i.e., the targets of hate, from reinforcing hateful, dehumanising or derogatory statements. The authors state they do not support or endorse any of the examples of hateful language presented in this paper. There is a risk of harm to the authors and to annotators who interact with, create and label hateful statements. The paper describes the steps taken to protect annotator well-being such as regular check-ins and provision of mental health support and resources. There is a risk of harm to readers and reviewers from the verbatim quoted examples used to illustrate the composition of \textsc{HatemojiCheck}. Some examples are needed to demonstrate the problem of emoji-based hate but steps have been taken to warn of and reduce the risk of harm.
\item \textbf{Preview} The work includes a content warning directly after the abstract, at least a page before any harmful content is displayed. The content warning is in red to maximise visibility. They also include a section-specific content warning before Section 2.2, which includes a number of hateful examples (see Figure \ref{fig:hmoji_sec_warning}).
\item \textbf{Distance} Only a minimal amount of examples is included to illustrate \textsc{HatemojiCheck}'s composition. All verbatim quotes in Section 2 are coloured in grey text to give the examples less prominence. All other examples in the paper (e.g., in Table 4) are presented with placeholders for the [IDENTITY]. 
\item \textbf{Disclaim} All examples are taken from \textsc{HatemojiCheck} which is a synthetically-generated dataset. The origins of hateful examples is made clear in the content warning on page 1.
\item \textbf{Respect}  For slurs and profanities in text, the authors star out the first vowel with an asterisk (Table 5). However, using an asterisk to star out emoji is not possible as it would obscure the entire character. Annotator demographics are described in the paper's appendix, and the authors sought an annotation team that prioritised diverse perspectives. All quoted examples are synthetic so contain no personally-identifying information. 
\end{enumerate}

\subsection{Misinformation}

\citet{zubiaga2016analysing} analysis misinformation in detail, providing an annotation scheme and dataset for rumour veracity and the reactions that people take toward rumours on the web.

\begin{enumerate}
\item \textbf{Risk of harm protocol}  There is a risk of harm to the authors of comments by leaving their names in the examples or data and thus removing their right to be forgotten. There's a risk of harm to paper authors being misquoted about misinformative claims. There's a risk of harm to readers (and users of the accompanying dataset) in exposure to false ideas or reports. 

\item \textbf{Preview} The work is clearly about rumours, which are mentioned in the title and multiple times on the first page. An explicit statement could be added that the examples in this paper are false, and do not reflect author views.

\item \textbf{Distance} The examples and claims in the paper are presented as rumours or of questionable veracity, which is stated in the accompanying caption. Examples are presented in a visually distinct way. In some cases the example is fictitious and abstract, thus minimising the amount of material included.

\item \textbf{Disclaim} The work doesn't explicitly disclaim the rumours in situ, but does enumerate the true narratives relative to each story and state which claims are false.

\item \textbf{Respect} While some rumour texts are included, usernames are removed in the paper through blurring or replacement with placeholders.

\begin{figure}
    \centering
    \includegraphics[width=\columnwidth, frame]{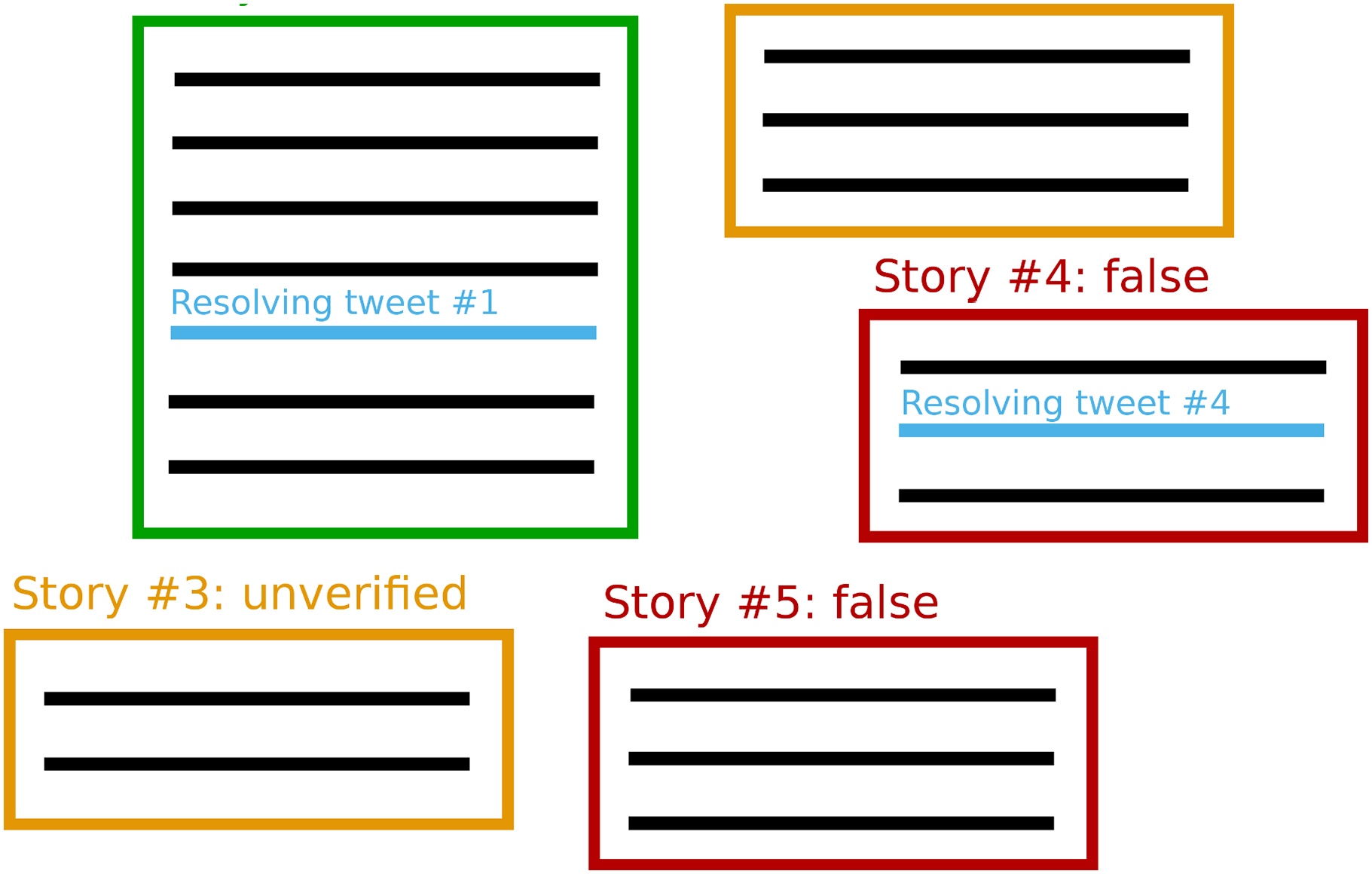}
    \caption{Figure from~\citet{zubiaga2016analysing} where the authors have minimised the amount of misinformative content needed to convey their point by using an abstraction instead of verbatim content.}
    \label{fig:misinfo_ex}
\end{figure}

\end{enumerate}

\end{document}